\documentclass[cameraready]{Interspeech}

\title{Do Speech Emphasis Models Generalize across Languages and Emotions?}

\author[affiliation={1,2}, orcid=0000-0002-1486-6409]{Megan}{Wei}
\author[affiliation={1}, orcid=0000-0001-9610-5648]{Deepali}{Aneja}
\author[affiliation={1}]{Jiaqi}{Su}
\author[affiliation={1}, orcid=0009-0007-9593-4971]{Yunyun}{Wang}
\author[affiliation={1}]{Haonan}{Chen}
\author[affiliation={1}, orcid=0000-0003-0161-5915]{Zeyu}{Jin}

\address{
    $^1$ Adobe Research, USA \\
    $^2$ Brown University, USA
}

\email{meganwei@brown.edu, aneja@adobe.com, jsu@adobe.com, yunyunw@adobe.com, haonanc@adobe.com, zejin@adobe.com}


\usepackage{comment}

\usepackage{xcolor}

\begin{document}

\maketitle


\begin{abstract}

Prosodic emphasis varies across languages, emotions, and speaking styles, yet existing emphasis detection models are largely trained and evaluated on monolingual neutral read speech. We introduce MMEE (Multilingual Multi-Emotion Emphasis), a corpus of 10,000 professionally recorded expressive utterances (14.13 hours) across 7 languages and 34 emotion/style categories, with three-level perceptual labels (10 annotations per sample). We benchmark two state-of-the-art architectures under monolingual, cross-lingual, multilingual, cross-emotion, cross-dataset, and data-scale settings. Monolingual models show limited zero-shot transfer, degrading across typologically distant languages, while multilingual training substantially improves robustness. Models transfer robustly between high- and low-arousal emotions; bidirectional transfer between synthetic and perceptual benchmarks suggests shared prosodic structure; and performance stays robust even at smaller training scales.

\end{abstract}

\section{Introduction}

Prosodic emphasis plays a key role in spoken communication, signaling contrast, focus, speaker intent, and affect. 
For example, the same words can convey different meanings depending on emphasis: “You can’t sit \emph{here}” can reject a location, while “You can’t \emph{sit} here” can reject the action itself.
Accurate modeling of emphasis is essential for expressive text-to-speech and prosody control \cite{seshadri2022emphasis,Suni2020,roekhaut10_speechprosody,joly23_ssw,liu2024emphasis,chien24b_interspeech,oh2024diffprosody,eme-tts,BauerZMD25_emphasiscontrol_SSW}, speech-to-speech translation \cite{deseyssel2023emphassess}, and reasoning \cite{yosha2025stresstest} about user intent. Despite many efforts toward modeling speech prosody \cite{vaidya,morrison2024crowdsourced,deseyssel2023emphassess,de2023prosaudit,yosha2025whistress,hung-etal-2025-exploring}, it remains underexplored in multilingual and emotionally expressive settings. As we deploy these systems globally, it is critical to build robust, generalizable emphasis detection models that can accurately capture emphasis cues across languages, cultures, and emotional states.

There has been significant work towards developing emphasis models and benchmarks. Morrison et al. \cite{morrison2024crowdsourced} crowdsourced perceptual emphasis annotations on LibriTTS \cite{zen2019libritts} read speech from audiobooks and trained acoustic models for prominence prediction. EmphAssess \cite{deseyssel2023emphassess} benchmarks emphasis transfer in speech-to-speech systems, using synthetic speech data with prescribed emphasis. WhiStress \cite{yosha2025whistress} augments a frozen Whisper model with a token-level stress-detection head, trained on synthetic speech and LLM-generated emphasis labels. Earlier expert-annotated corpora such as Aix-MARSEC \cite{Auran2004TheAP} encode structural stress patterns via narrow rhythm unit (NRU) notation. 

However, existing approaches share several limitations: they are predominantly English-only, often rely on synthetic or prescribed emphasis rather than human perceptual judgments, and have a limited range of speaking styles. Moreover, emphasis is typically treated as a binary classification task, despite its inherently graded nature. Thus, it remains unclear whether emphasis detection models learn language-specific prosodic patterns and generalize across different emotions and speaking styles, where pitch or duration patterns for emphasis vary (e.g. higher vs lower arousal emotions).

To address these gaps, we conduct a large-scale multilingual study using MMEE, a curated expressive speech corpus of 10,000 samples across 7 macro-languages, 10 regional varieties, and 34 emotions and speaking styles, with human-annotated, graded, word-level emphasis scores. We ask: (1) how well do emphasis detectors transfer across languages and language families? (2) does multilingual training improve robustness over monolingual training? (3) do models trained on one arousal regime generalize to another? and (4) do human-perceptual and synthetic emphasis labels support transferable representations? Using MMEE, we benchmark two state-of-the-art speech emphasis detection models, EmphaClass \cite{deseyssel2023emphassess} and WhiStress \cite{yosha2025whistress}, across the following settings: monolingual, cross-lingual, multilingual, cross-emotion arousal, cross-dataset generalization, and training dataset scale.\footnote{Website: \url{https://multilingual-speech-emphasis.github.io}}

\begin{figure}[t]
  \centering
  \includegraphics[width=\linewidth]{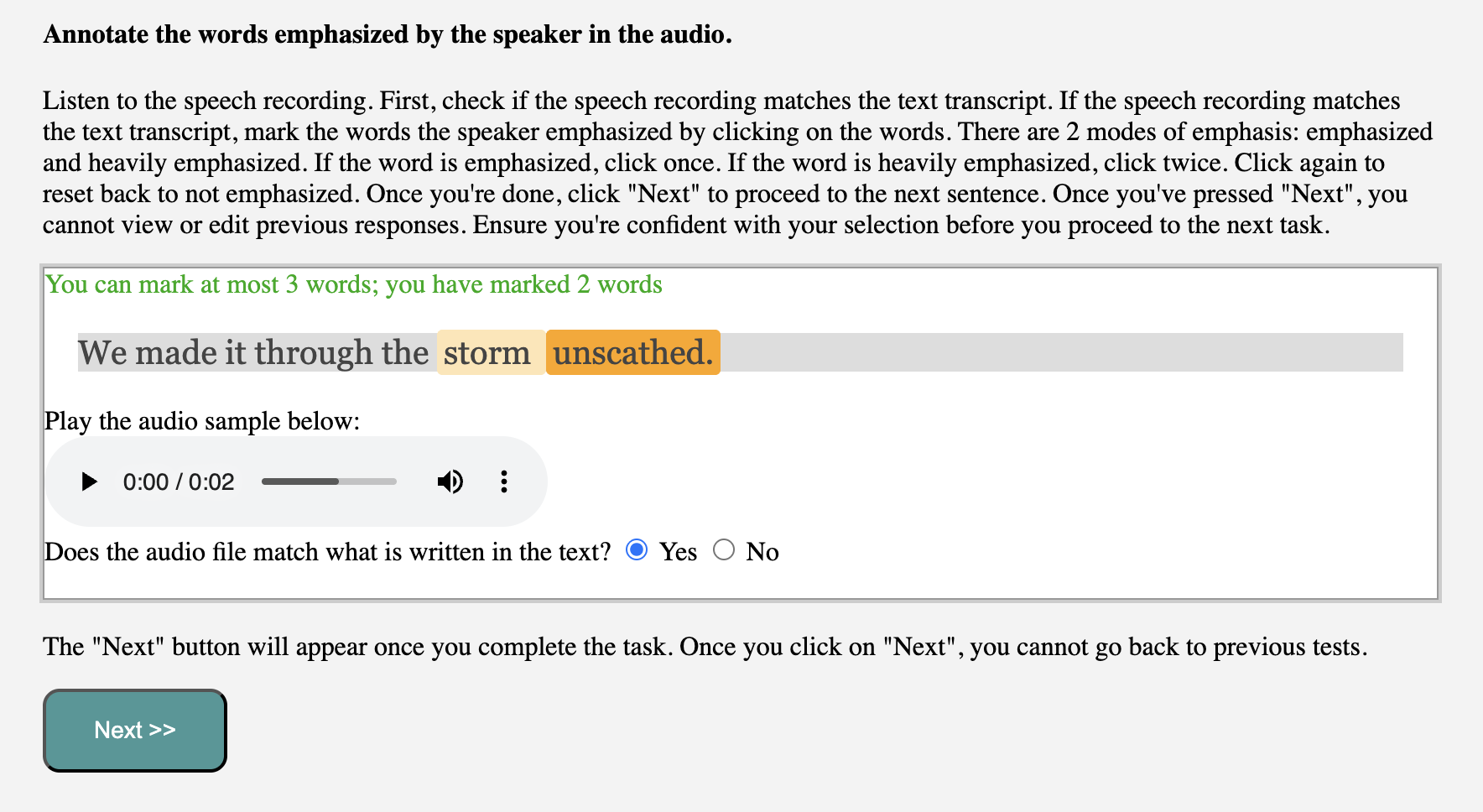}
  \vspace{-1.4em}
  \caption{Speech emphasis annotation interface on Prolific. Participants click on words in the transcript they perceive as emphasized after listening to the audio.}
  \label{fig:prolific}
   \vspace{-1.4em}
\end{figure}

\begin{table*}[t]
\small
\caption{Comparison of emphasis datasets. Label source: Human = listeners mark perceived emphasis; LLM = GPT-4o-mini marks emphasis; Prescribed = script specifies emphasis, TTS synthesizes; Expert = linguists apply NRU.}
\label{tab:dataset-comparison}
\centering
\begin{tabular}{l|ccccc}
\toprule
 & \textbf{Ours} & Morrison \cite{morrison2024crowdsourced} & TinyStress-15K \cite{yosha2025whistress} & EmphAssess \cite{deseyssel2023emphassess} & Aix-MARSEC \cite{Auran2004TheAP} \\
\midrule
Languages & 7 (+3) & 1 & 1 & 1 & 1 \\
Speech Source & Voice actors & Read (LibriTTS) & Synthetic & Synthetic & Broadcast (BBC) \\
Utterances & 10,000 & 3,626 & 16,000 & 3,652 & 2,400 \\
Hours & 14.13 & 6.42 & 16.03 & 2.42 & 5.65 \\
Speakers & 202 & 18 & 10 & 4 & 53 \\
Ann./sample & 10 & 1--8 & 1 & 1 & 1 \\
Label Source & Human (Prolific) & Human (MTurk) & LLM & Prescribed (script) & Expert (NRU) \\
Granularity & 3-level → scalar & Binary → scalar & Binary & Binary & Binary (NRU) \\
Emotions \& Styles & 34 & -- & -- & -- & -- \\
Cohen's $\kappa$ & 0.285--0.518 & 0.226 & N/A & N/A & N/A \\
\bottomrule
\end{tabular}
\end{table*}

\begin{figure*}[t]
  \centering
  \includegraphics[width=\textwidth]{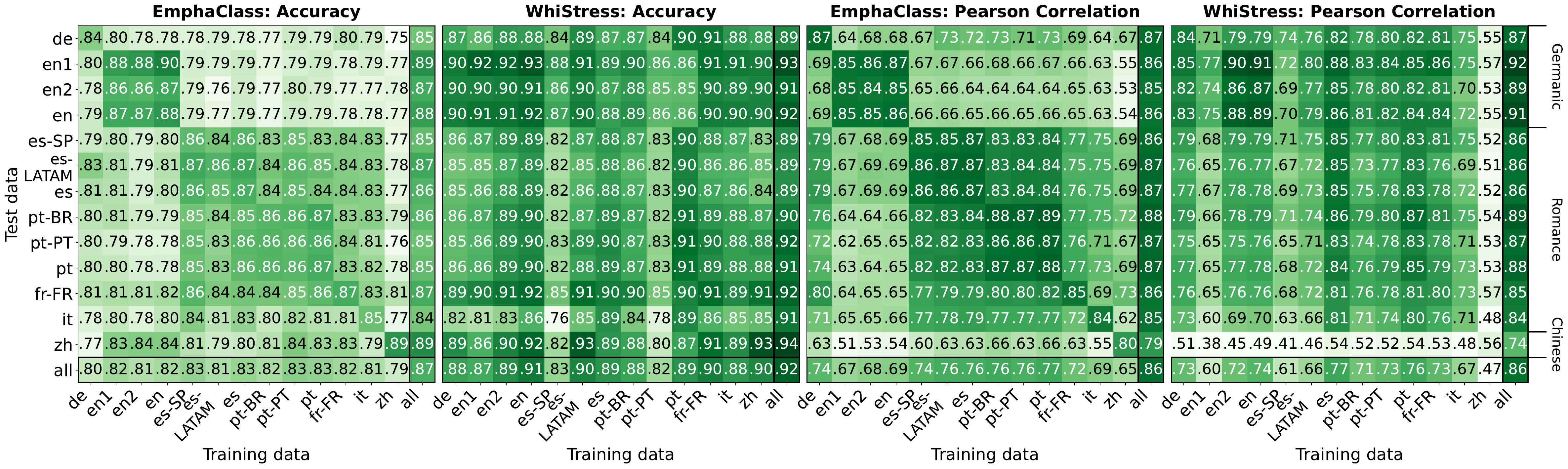}
  \caption{Binary Accuracy and Scalar (Pearson Correlation) results for EmphaClass and WhiStress. The ``all" dataset is the full multilingual set (combined test sets of each of the 10 regional varieties). ``en" represents ``en1" (English Americas) and ``en2" (English Other) combined; ``es" represents ``es-SP" and ``es-LATAM" combined; and ``pt" represents ``pt-BR" and ``pt-PT" combined.}
  \label{fig:heatmap-combined}
\end{figure*}
\section{Dataset}

\subsection{Speech Corpus}
We leverage a proprietary multilingual expressive speech corpus we collected internally as the foundation for annotating emphasis and conducting analyses. The dataset spans 34 categories of emotions and speaking styles across various languages and dialectal varieties. 
Scripts and accompanying performance instructions are generated with the assistance of an LLM for each language and style category.
The scripts are designed to naturally induce prosodic variation, while the instructions describe the target emotion/style and reference emphasis placements. In practice, voice talents often produce their own interpretations of the scripts, yielding diverse emphasis realizations using prosodic cues like increased intensity, higher pitch, and elongated duration.

Native voice talents are recruited and screened through an audition process that evaluates their performance with respect to naturalness, expressiveness, and alignment with the intended emotional styles.
Each qualifying voice talent performs a set of scripts covering all emotions and speaking styles, with recordings reviewed for acoustic quality (low background noise, limited reverberation, and no audible distortion).
The resulting corpus comprises 10 regional categories across 7 macro-languages, with approximately 20 speakers per accent: English Americas (North American, Southern, African American), English Other (Indian, Australian, British), Spanish (Spain), Spanish (Latin America), Portuguese (Portugal), Portuguese (Brazil), German, French, Italian, and Mandarin Chinese. 

\subsection{Data Curation}
We curate 1,000 high-quality utterances per language class (10,000 total, 14.13 hours) from the speech corpus using a multi-stage pre-processing pipeline to produce clean, accurately segmented clips for fine-grained emphasis annotation.

The raw recordings undergo a uniform background noise reduction pass. We obtain word-level timestamps and transcripts with Qwen3-ASR~\cite{Qwen3-ASR}, split recordings into 1--2 sentence utterances, and refine boundaries at low-energy valleys of the RMS energy envelope to avoid truncated phonemes or bleed-in. Each clip is re-transcribed with Qwen3-ASR~\cite{Qwen3-ASR} and compared to the source script using normalized sequence similarity (threshold $\geq$99\%). Additional signal-level checks are applied for abrupt waveform boundaries, excessive leading/trailing silence ($>$1 s), and abnormally short word durations. A trim sensitivity test (trimming 200 ms from each end and re-measuring similarity) flags overly tight boundaries. Clips failing any check are iteratively recropped and re-validated.
Flagged cases (e.g. bleed-in or boundary inaccuracies) are adjusted via voice activity detection (Silero-VAD)~\cite{SileroVAD} to locate precise speech onset and offset, followed by re-validation.

To filter out voice performance artifacts, we use GPT-5.2 as a judge \cite{zheng2023judging} comparing each transcript against its source script, requiring unanimous acceptance across three independent runs to mitigate hallucination risks. In Section \ref{sec:emphannotation}, the human-judged audio-transcript match filter provides an additional safeguard.
Furthermore, to create a diverse, robust dataset, we remove duplicate utterances per language based on their scripts \cite{lee2021deduplicating}, and balance the emotion distribution to approximately 29-30 samples per emotion per 1,000-sample language class. 

\subsection{Emphasis Annotation}\label{sec:emphannotation}
To obtain word-level emphasis, we solicit human annotators on Prolific. In our interface in Figure \ref{fig:prolific}, annotators listen to the audio and mark the words perceived as emphasized, on a three-level scale: not emphasized, emphasized, or heavily emphasized. This graded scheme captures nuances of perceived emphasis beyond a binary distinction.

Defining emphasis as a perceptual judgment by native listeners \cite{morrison2024crowdsourced}, we crowdsource emphasis annotations from fluent native speakers instead of trained linguists. Annotator prerequisites include: no hearing or literacy difficulties; the study language as their primary and fluent language; an undergraduate degree or higher; a Prolific approval rate of 99-100\%; and at least 25 prior completed studies. We request 500 task slots per language (20 audio samples per task), targeting 10 annotations per sample. Annotators are compensated \$15/hour. 

Our Prolific interface includes detailed instructions and an example. We include the following quality checks: annotators are required to (1) listen to each audio clip at least once, (2) select at least one emphasized word per sample, and (3) mark no more than 30\% of the words in an utterance as emphasized. All submissions are final, with no option to revise earlier responses. 

Annotators can flag whether the audio matched the displayed transcript. Among 10,000 samples, 78 (0.78\%) have $\geq$2 mismatch reports; after systematic Qwen3-ASR~\cite{Qwen3-ASR} and GPT-5.2 review, we recrop 8 clips with genuine cropping issues (truncated syllables, bleed-in, or extraneous sounds), while the remaining 70 are confirmed valid.

Due to occasional annotator dropout, we dynamically launch reruns targeting under-annotated samples. For samples with more than 10 annotations, we randomly subsample to exactly 10 annotations with a fixed seed for reproducibility. In the following experiments, we aggregate the annotations in two modes: binary and scalar. In binary mode, if more than half of the 10 annotators mark a word as emphasized, the word is deemed emphasized. In scalar mode, we use the mean of the per-annotator ordinal scores (0 = not emphasized, 0.5 = emphasized, 1 = heavily emphasized), yielding a continuous prominence score richer than a binary label alone; averaging across 10 annotators reduces sensitivity to any single listener's perception.

\subsection{Dataset Comparison}
Table~\ref{tab:dataset-comparison} compares our dataset to existing emphasis datasets. Prior work falls into three paradigms. (1) \textbf{Synthetic TTS with prescribed or LLM-generated labels}: EmphAssess \cite{deseyssel2023emphassess} uses transcripts with prescribed emphasis markers; the TTS model is instructed to emphasize those words, so labels come from the script before synthesis, not from listening. TinyStress-15K \cite{yosha2025whistress} uses GPT-4o-mini to select stressed words, then Google TTS synthesizes with SSML prosodic adjustments; labels are LLM-generated, not human. (2) \textbf{Crowdsourced perceptual annotation}: Morrison et al.\ \cite{morrison2024crowdsourced} use MTurk workers who listen to LibriTTS read speech and click emphasized words; labels are human perceptual judgments, as in our dataset. (3) \textbf{Expert linguistic annotation}: Aix-MARSEC \cite{Auran2004TheAP, LEE201729} uses Jassem's NRU (narrow rhythm unit) notation; expert linguists mark prosodic structure; a word is stressed if it contains the first syllable of an NRU. This is phonological (structural), not perceptual. 

Our dataset is, to our knowledge, the first to combine (i) \textbf{multilingual coverage} (7 macro-languages, 10 varieties), (ii) \textbf{expressive emotional speech} across 34 categories, and (iii) \textbf{graded human emphasis annotations} (3-level) with a balanced 10 annotations per sample design.

We report multiple agreement metrics in our dataset. When collapsing the three levels to binary (emphasized vs not), average pairwise Cohen's $\kappa$ ranges from 0.285 (Chinese, fair) to 0.518 (Portuguese Brazil, moderate), with a pooled value of 0.451 (moderate) -- substantially higher than Morrison et al.'s Cohen's $\kappa$ of 0.226 (fair) on LibriTTS. Fleiss' $\kappa$ (pooled 0.446, 95\% CI [0.442, 0.449]) yields similar values. Krippendorff's $\alpha$ (ordinal, 3-level) is 0.461 (95\% CI [0.457, 0.465]) on the pooled dataset, supporting the validity of the 3-level scheme. Chinese shows lower agreement, likely due to its tonal system and different prosodic cues for emphasis. Emphasis rate (percentage of emphasized words) varies by language (15--22\%) and is relatively stable across emotions (17--22\%), indicating consistent emphasis elicitation from scripts.

\section{Methods}
We benchmark two state-of-the-art models, EmphaClass \cite{deseyssel2023emphassess} and WhiStress \cite{yosha2025whistress}, on MMEE, using a fixed 80/10/10 train/validation/test split, shared across models. All experiments are conducted on 8 NVIDIA 80 GB A100s.

EmphaClass \cite{deseyssel2023emphassess} finetunes a 1B-parameter multilingual SSL model (XLS-R) \cite{xlsr} built on Wav2Vec 2.0 \cite{wav2vec2} for frame-level binary classification. A word is emphasized if $>50\%$ of the frames are classified as emphasized. We extend it to scalar regression, by swapping the classification head with a regression head (linear + sigmoid) trained with MSE loss. For variable-length sequences, we switch the model's original zero-padding to $-100$, so padded positions are excluded from the loss and not conflated with the ``not emphasized" class. We train for 15 epochs with learning rate $7.97\times10^{-5}$, 12.5\% warmup, batch size 8 (4 for multilingual/arousal) and gradient accumulation 3.

WhiStress \cite{yosha2025whistress} consists of a frozen Whisper \cite{radford2022whisper} encoder-decoder, an additional decoder block, and an FCNN classifier head producing per-token emphasis scores. The original version of WhiStress uses the \texttt{whisper-small.en} checkpoint. To support multilingual processing, we use the \texttt{whisper-small} checkpoint and add language conditioning to Whisper, passing the language token for decoding. Hidden states from encoder and decoder layer $9$ are combined in the additional decoder block, which is then passed into the classifier head. Training uses 2 epochs, learning rate $5\times10^{-4}$, 5\% warmup, weight decay 0.01, batch size 32. Binary mode uses weighted cross-entropy loss ([1, 2.33]); scalar mode uses BCE loss.

\begin{figure}[t]
    \centering
    \includegraphics[width=\linewidth]{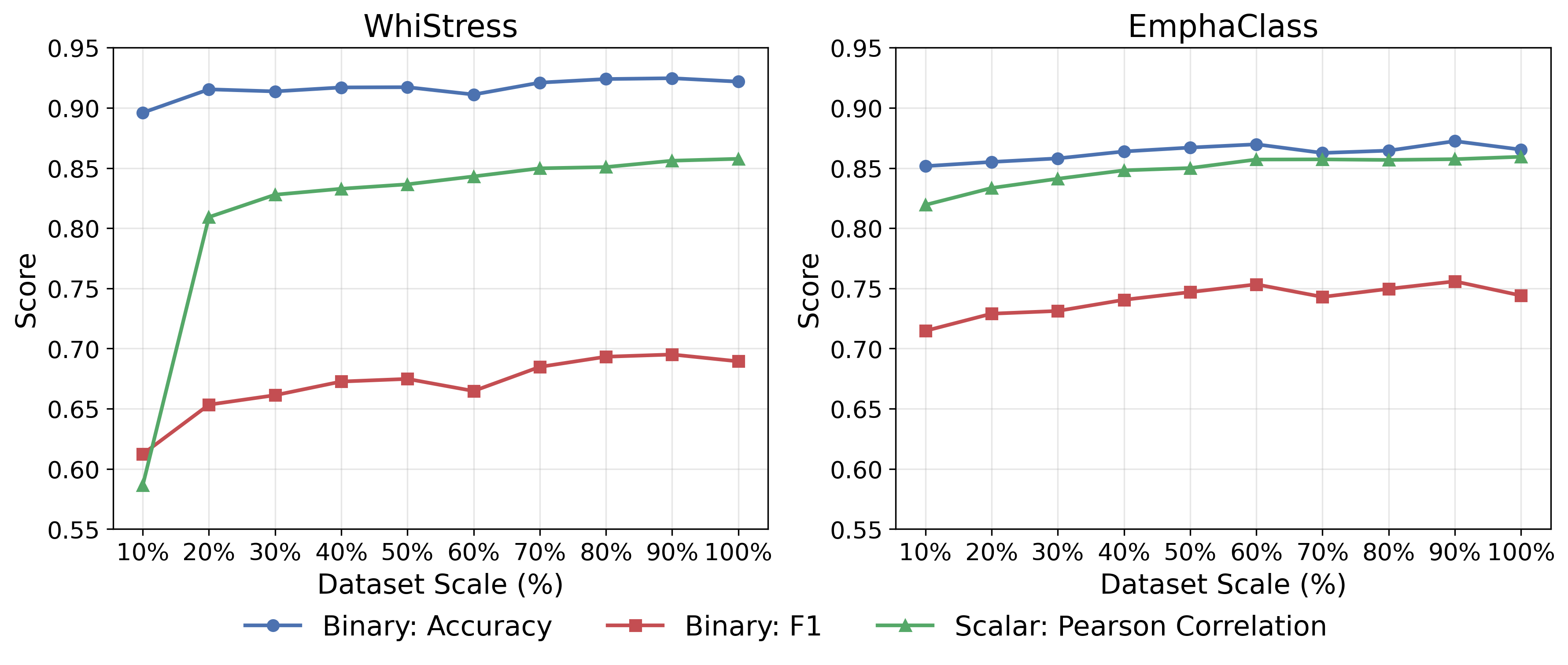}
    \caption{
    Binary (Accuracy, F1) and scalar (Pearson Correlation) performance as a function of training data scale.
    }
    \label{fig:dataset_scale}
\end{figure}

\begin{table}[t]
\centering
\caption{High/low arousal evaluation (Binary Accuracy and Scalar Pearson Correlation).}
\label{tab:arousal}
\vspace{-0.5em}

\begin{tabular}{lcccc}
\toprule
Condition & \multicolumn{2}{c}{EmphaClass} & \multicolumn{2}{c}{WhiStress} \\
\cmidrule(lr){2-3} \cmidrule(lr){4-5}
& Acc & Pearson & Acc & Pearson \\
\midrule
High $\rightarrow$ High & 0.848 & 0.846 & 0.918 & 0.833 \\
Low $\rightarrow$ Low & 0.871 & 0.840 & 0.912 & 0.823 \\
High $\rightarrow$ Low & 0.857 & 0.814 & 0.908 & 0.819 \\
Low $\rightarrow$ High & 0.857 & 0.833 & 0.920 & 0.814 \\
\bottomrule
\end{tabular}
\end{table}
\vspace{-0.5em}

\begin{table}[t]
\centering
\caption{Cross-dataset generalization (Binary Accuracy).}
\vspace{-0.4em}
\label{tab:cross-dataset}
\begin{tabular}{l c}
\toprule
\textbf{Direction} & \textbf{Acc} \\
\midrule
\addlinespace
\multicolumn{2}{l}{\textit{EmphaClass}} \\
MMEE (en) $\rightarrow$ EmphAssess & 0.886 \\
MMEE (all) $\rightarrow$ EmphAssess & 0.875 \\
EmphAssess $\rightarrow$ MMEE (en) & 0.798 \\
\addlinespace
\multicolumn{2}{l}{\textit{WhiStress}} \\
MMEE (en) $\rightarrow$ TinyStress-15K & 0.873 \\
MMEE (all) $\rightarrow$ TinyStress-15K & 0.876 \\
TinyStress-15K $\rightarrow$ MMEE (en) & 0.881 \\
\bottomrule
\end{tabular}
\end{table}
\section{Results and Discussion}

\subsection{Monolingual, Cross-Lingual, and Multilingual}
Monolingual experiments are run on all language dialects and classes, across 13 configurations. Cross-lingual experiments involve training on one language and testing on each of the other languages. Multilingual (all) trains on the full dataset (8,000 training samples) and tests on individual languages and the reserved all test set. We also test generalization of monolingual-trained models to multilingual. We use a fixed train/validation/test split for each language; ``all'' combines the corresponding train/validation/test split from monolingual sets.

Figure~\ref{fig:heatmap-combined} shows that monolingual models achieve strong in-language performance across most languages. Chinese consistently underperforms, consistent with its tonal prosody, in which F0 simultaneously encodes lexical tone and prominence \cite{XU199955}, and lower inter-annotator agreement.

Zero-shot cross-lingual transfer degrades with typological distance. Within family transfer (e.g., Romance-Romance) approaches monolingual performance, while transfer between Romance/Germanic and Chinese drops substantially. 

Multilingual pooled training (``all'') demonstrates strong cross-lingual robustness, often matching or exceeding monolingual training performance, suggesting that exposure to diverse prosodic patterns strengthens emphasis representations and mitigates overfitting to language-specific cues.

\subsection{Data Scale}
We vary the training data scale from 10\% to 100\% (800 to 8,000 utterances), with even distribution across the 10 language varieties. The subsets are nested (e.g. the 10\% set is fully included in 20\% set) drawn from per-language shuffled utterance lists. Validation and test sets are fixed at 1,000 samples each.

Figure \ref{fig:dataset_scale} shows performance as a function of training data scale. Both models benefit from increased training data, with initial rapid gains and diminishing returns thereafter. EmphaClass maintains strong Binary Accuracy and Pearson Correlation across all scales, while WhiStress exhibits weaker scalar performance at 10\%, but  improves at 20\%. These results suggest these models are relatively data-efficient and benefit from multilingual data diversity, which has practical implications for extending emphasis modeling to new language families.

\subsection{Arousal}
We construct high- and low-arousal subsets following the arousal dimension of the circumplex model \cite{russell1980circumplex}. High arousal comprises excitement, happiness, pride, determination, anger, fear, anxiety, frustration, and disgust; low arousal comprises calmness, relief, love/affection, hopefulness, sadness, boredom, shame, embarrassment, and contempt. The high-arousal training set contains 2,070 samples with 270 each for validation and test, balanced across languages and emotions; likewise for low arousal.

Arousal affects the way emphasis is realized \cite{SCHERER2003227}: high-arousal speech tends toward higher and more variable pitch \cite{paeschke1999f0, bussof0}, greater intensity, and faster tempo, whereas low-arousal speech tends toward lower, flatter pitch and slower, lengthened delivery \cite{frick, murray1993toward}. We test each model in-domain and across the arousal boundary. Despite these acoustic differences, both models perform strongly in-domain and transfer robustly across arousal conditions in both binary and scalar tasks (Table~\ref{tab:arousal}), suggesting that the emphasis signal they exploit is partially separable from arousal-driven acoustic variation.

\subsection{Cross-Dataset Generalization}
We evaluate cross-dataset generalization for EmphaClass and WhiStress. First, we test if the model trained on MMEE (English-only and all languages) generalizes to the EmphAssess or the TinyStress-15K dataset, using their provided test partition. Then, we test the original EmphaClass and WhiStress checkpoints on MMEE (English). We did not test the original EmphaClass and WhiStress models on MMEE (all languages), due to lack of support in multilingual tokenization. All evaluations use binary mode, consistent with the binary label format 
of these datasets.

Table \ref{tab:cross-dataset} reveals strong bidirectional transfer. In EmphaClass, MMEE (en) $\rightarrow$ EmphAssess substantially outperforms the reverse direction. Meanwhile, TinyStress-15K $\rightarrow$ MMEE (en) marginally outperforms the reverse direction, likely due to stronger English acoustic-prosodic grounding in the \texttt{whisper-small.en} checkpoint used by the original WhiStress, compared to the \texttt{whisper-small} version used in the MMEE-trained model. These results suggest that synthetic and human-perceptual emphasis annotations share substantial prosodic signal, despite differences in label and data sources.

\section{Conclusion}
We investigate whether pretrained speech models learn universal representations of prosodic emphasis through a large-scale multilingual study using MMEE. Emphasis representations are partially universal, but fracture at typologically distant languages. Cross-lingual transfer follows language family structure and drops most strongly for Mandarin Chinese, while multilingual pooled training is more robust. Models generalize well across arousal conditions despite different acoustic cues. Strong bidirectional transfer between human-perceptual and synthetic benchmarks reveals that emphasis is robust to annotation paradigm and data source. Furthermore, data-scale gains concentrate in the first few thousand samples, lowering the barrier to emphasis modeling for low-resource languages.

\section{Generative AI Use Disclosure}
Cursor and Claude Code were used to assist experiment implementation. The authors thoroughly reviewed and validated all outputs throughout the implementation and experimentation process. Claude was used for minor editing of the draft (e.g. reducing word count, formatting tables); the original draft and all research contributions are the authors'. The authors used an LLM-as-a-Judge (GPT-5.2) as one proxy for quality checks on transcripts during dataset curation; signal-based and human-based evaluations were also used to confirm transcript quality of the dataset.

\bibliographystyle{IEEEtran}
\bibliography{mybib}

\end{document}